\documentclass[letterpaper, 10 pt, conference]{ieeeconf}
\usepackage{hyperref}

\IEEEoverridecommandlockouts
\usepackage{cite}
\usepackage{graphicx} 
\usepackage{mathptmx}
\usepackage{times}
\usepackage{amsmath}
\usepackage{amssymb}
\usepackage{txfonts}
\usepackage{caption}
\usepackage{color}
\usepackage[table,xcdraw]{xcolor}
\usepackage{booktabs}
\usepackage{overpic}
\usepackage{bm}
\usepackage{multirow}
\usepackage{siunitx}
\usepackage[capitalise, nameinlink]{cleveref}

\title{\LARGE \bf
Maintaining Plasticity in Reinforcement Learning: A Cost-Aware Framework for Aerial Robot Control in Non-stationary Environments
}

\author{Ali Tahir Karasahin$^{1,2}$, Ziniu Wu$^{2}$, and Basaran Bahadir Kocer$^{2}$
\thanks{*Ali Tahir Karasahin was supported by the Turkish Scientific and Technological Research Council (Grant Ref. 1059B192302726). Ziniu Wu was supported by the Engineering and Physical Sciences Research Council (Grant Ref. EP/W524414/1).}
\thanks{$^{1}$Ali Tahir Karasahin is with Faculty of Engineering, Department of Mechatronics Engineering, Necmettin Erbakan University, Turkey.}%
\thanks{$^{2}$Authors are with the School of Civil, Aerospace and Design Engineering, University of Bristol, UK.}
}

\begin{document}

\maketitle
\thispagestyle{empty}
\pagestyle{empty}

\begin{abstract}

Reinforcement learning (RL) has demonstrated the ability to maintain the plasticity of the policy throughout short-term training in aerial robot control. However, these policies have been shown to loss of plasticity when extended to long-term learning in non-stationary environments. For example, the standard proximal policy optimization (PPO) policy is observed to collapse in long-term training settings and lead to significant control performance degradation. To address this problem, this work proposes a cost-aware framework that uses a retrospective cost mechanism (RECOM) to balance rewards and losses in RL training with a non-stationary environment. Using a cost gradient relation between rewards and losses, our framework dynamically updates the learning rate to actively train the control policy in a variable wind environment. Our experimental results show that our framework learned a policy for the hovering task without policy collapse in variable wind conditions and has a successful result of 11.29\% less dormant units than L2 regularization with PPO. Project website: \url{https://aerialroboticsgroup.github.io/rl-plasticity-project/}

\end{abstract}

\section{Introduction}

Over the past decade, aerial robots have been applied to many domains, including environmental monitoring and forest ecology \cite{kocer2021forest,ho2022vision,lan2024aerial,bates2025leaf}. To further broaden their usability in wild environments, it is crucial to develop active learning capabilities that enable controllers to adapt in terms of unmodeled dynamics and external disturbances for applications where the conditions are dynamically changing. One of the desirable approaches is using reinforcement learning (RL) which enables robots to autonomously explore an optimal policy based on trial-and-error interactions with its environment \cite{13_IJRR_RL_Survey}. 

RL can directly learn the value function or the policy without any explicit modeling of the transition dynamics. In \cite{24_RAL_learningtoflyinseconds}, an RL-based training methodology is introduced for hovering control and trajectory tracking of an aerial robot. They show that integrating curriculum learning with a highly optimized simulation environment enhances sample efficiency and accelerates the training process. Remarkably, an autonomous aerial robot using a deep reinforcement learning policy can compete against the human world champions in real-world drone racing \cite{kaufmann_champion-level_2023}. To enable RL agents to learn in non-stationary environments, the learning framework itself must maintain long-term learning capabilities without policy collapse. While most current applications and methods in RL emphasize stability to ensure that learned policies remain robust under fixed conditions, they often under-explore plasticity. This oversight is particularly problematic in non-stationary settings, where environmental dynamics such as variable wind disturbances can make a previously robust policy ineffective \cite{dohare_loss_2024}. Therefore, mitigating plasticity loss in RL training frameworks is critical in non-stationary environment scenarios to make the aerial robot adaptive to environmental changes or multiple tasks \cite{25_RAL_multitask_rl_yunlong}. 



   \begin{figure}[t]
      \centering
      \includegraphics[width=1\linewidth]{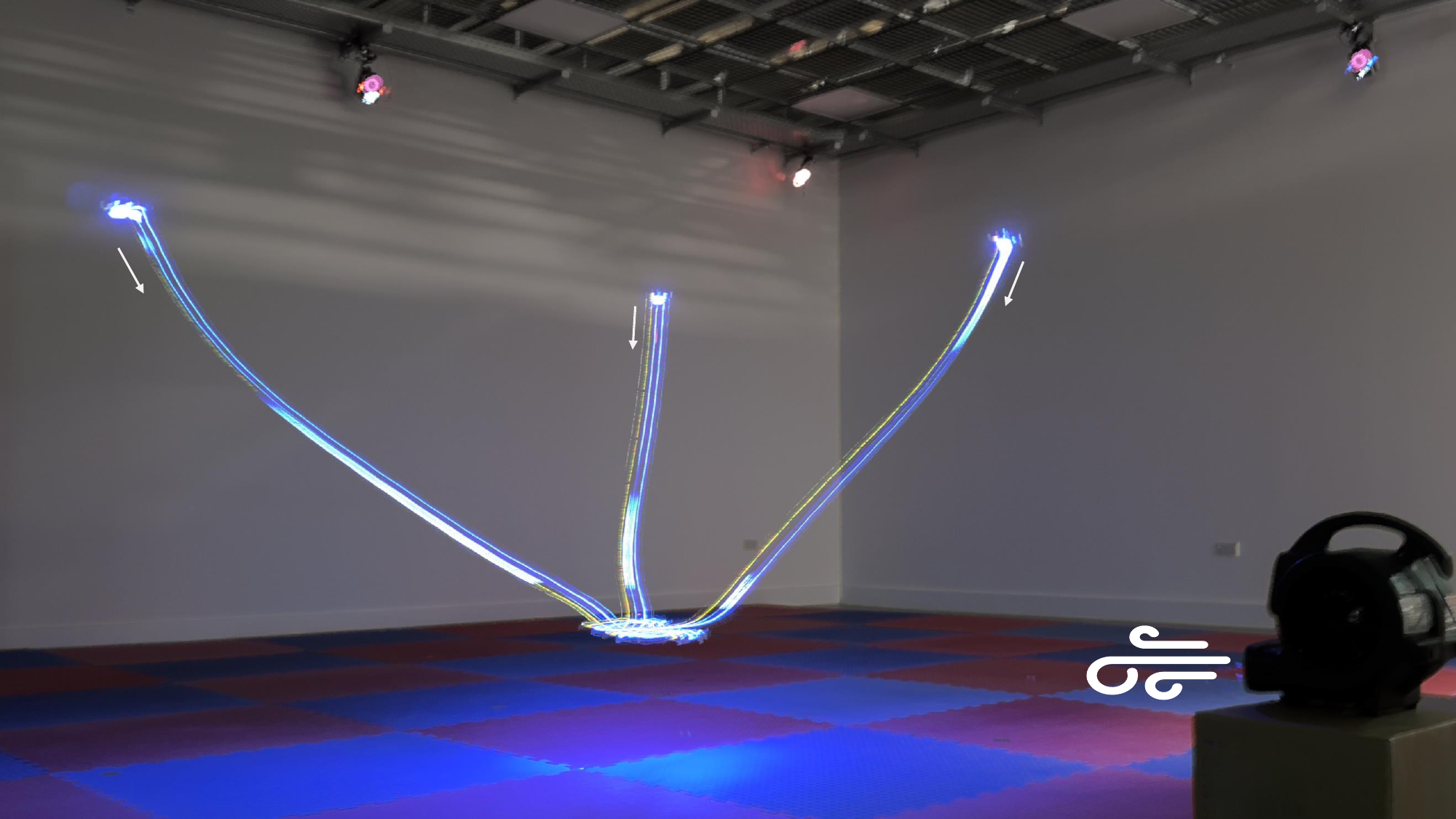}
      \caption{Example of aerial robot hovering from different initial positions under variable wind conditions.}
      \label{figurelabel}
   \end{figure}




To address the plasticity loss challenge, we explore the biologically motivated mechanisms to reshape learning dynamics. We can trace back the relationship between reward-prediction circuits and error processing systems to change the learning rate during long-term training. Therefore, in this paper, we show that a balance between exploration and exploitation can be achieved when rewards and losses are considered together from a retrospective cost perspective. This interaction is reflected in cognitive science as the development of goal-directed behaviour and adaptive decision-making, and in neuroscience as the relationship between reward-prediction circuits, also known as dopaminergic pathways and error processing systems. 
%
%
To the best of our knowledge, this is the first work to explicitly address plasticity loss in RL for aerial robot learning. Therefore, we have the following highlights in this paper:


\begin{itemize}
    \item We propose a cost-aware learning mechanism investigation for balancing rewards and losses in reinforcement learning, inspired by cognitive and neuroscientific principles. This is based on a dynamic adjustment that adapts exploration and exploitation strategies based on recent outcomes.
    \item We provide empirical evidence that shows the baseline Proximal Policy Optimization (PPO) policy collapses with dynamic wind changes for hover flight. To address this, we investigate L2 regularization and the retrospective cost mechanism and benchmark them in simulation under variable wind conditions. 
    \item We discuss the interplay between reward circuits and error-processing systems, linking biological insights with artificial learning models. 
\end{itemize}



\section{Related Work}
\subsection{Plasticity Loss in RL}


In recent years, RL has become a preferred technique as an artificial agent and demonstrated significant success by establishing a relationship between sensory input and actions in complex tasks including aerial perching \cite{bbk_learning_tether}. While significant progress has been made in RL, a critical challenge, loss of plasticity, arises in RL with non-stationary environment. In other words, the agent loses or reduces the ability to adapt to new experiences over time. An example of plasticity loss in RL long-term training is demonstrated in \cite{dohare_loss_2024}, where an ant-like agent was trained to adapt its locomotion strategy as ground friction varied. Using a standard PPO framework, the agent initially learned to adjust its gait as the friction changed. However, its ability to continually learn is reduced over long-time training. This phenomenon highlights how standard RL training methods fail to maintain plasticity in long-term learning scenarios. The study proposed that incorporating L2 regularization with continual backpropagation can mitigate plasticity loss. The agent retains its adaptive capability when environmental conditions shift.

To the best of our knowledge, there has been limited exploration of plasticity loss in RL long-term training within the field of aerial robotics. This research area remains an underexplored challenge in the domain.


\subsection{Cognitive Neuroscience Inspirations in RL}
In the context of RL, a series of studies inspired by neuroscience and cognitive science have been conducted. In \cite{nikishin_primacy_2022}, inspired by cognitive science, a relation to a primacy bias is established, which is a behaviour of relying on early interactions and ignoring the useful feature encountered later. It has been shown that a solution to this problem can be provided by periodically resetting the last layer of the network. 
According to the connections between neuroscience and RL, \cite{marblestone_toward_2016} proposes that the brain optimizes multiple cost functions to facilitate targeted learning, using efficient architectures tailored to specific cognitive tasks. It is thought that by combining both neuroscience and deep learning principles, the gap between biological and artificial intelligence can be closed according to this perspective \cite{hassabis_neuroscience-inspired_2017}. Therefore, we can take inspiration from the relationship between reward-prediction circuits and error-processing systems to migrate plasticity loss in RL. 



\section{Methodology}

\subsection{Aerial Robot Model}
To build an environment for policy training, we need a mathematical model that acts as a dynamic model of the aerial robot. In this study, we consider a 6-degree-of-freedom (DoF) system model for the quadrotor-type aerial robot. Let $ \bm{p} \in \mathbb{R}^3$, $\bm{v} \in \mathbb{R}^3$, and $g$ represent the position, linear velocity of the aerial robot, and gravity, respectively, given in the world frame. $\bm{e}_3$ is a basis vector $[0,0,1]^t$. Let $\bm{R} \in \mathrm{SO}(3)$ represent an attitude rotation matrix from the body to the world frame. Let $\bm{f}$, $m$, $\bm{J}$ and $\bm{d}$ represent the total thrust vector generated by the motors in the body frame, mass of the aerial robot, inertial matrix of the aerial robot and the translational disturbance force vector. Let $\boldsymbol{\omega} \in \mathbb{R}^3$, and $\boldsymbol{\tau}$ represent the angular velocity of the aerial robot and the torque vector in the body frame. $\bm{S}$ denotes the skew-symmetric matrix. The dynamic model of the aerial robot is denoted as:

\begin{equation} \label{drone_dynamics}
\begin{bmatrix}
\dot{\bm{p}}\\
m\dot{\bm{v}}\\
\dot{\bm{R}}\\
\dot{\boldsymbol{\omega}}
\end{bmatrix} = \begin{bmatrix}
\bm{v}\\
-mg\bm{e}_3 + \bm{R} (\bm{f} + \bm{d})\\
\bm{R}S(\boldsymbol{\omega})\\
\bm{J}^{-1}(\boldsymbol{\tau} - \boldsymbol{\omega} \times (\bm{J} \boldsymbol{\omega}))
\end{bmatrix}.    
\end{equation}

\subsection{RL Formulation}

To model the aerial robot's hovering task, we use \cref{drone_dynamics} to simulate its kinematics and dynamics. We define a Markov Decision Process (MDP) $M = (S, A, P, R, \gamma)$ where the $S$ represents the state $s \in S$ includes the position of the robot $[x, y, z]$, the linear velocity $[\dot{x}, \dot{y}, \dot{z}]$, the orientation $[\phi, \theta, \psi]$ and the angular velocity $[\dot{\phi}, \dot{\theta}, \dot{\psi}]$, $A$ is the action space, the action $\bm{a} \in A$ represents the control command input to the aerial robot. $P$ is the transition dynamics, $P(\bm{s}_{t+1}|\bm{s}_t, \bm{a}_t)$ is the probability of transitioning to the next state $\bm{s}_{t+1}$ given the actual state $\bm{s}_t$ and the output of the agent action $\bm{a}_t$. $R$ is the reward function, $r_t=R(\bm{s}_t, \bm{a}_t)$ designed to maintain the fixed position of the aerial robot in a hovering task. $\gamma$ is the discount factor that determines the importance of future rewards. In RL, the goal is the determined a policy $\pi_{\theta}(\bm{a}|\bm{s})$ parameterized by $\theta$ that maximizes the expected cumulative discounted reward $J(\pi_{\theta})$, the optimization objective is specified as:

\begin{equation}\label{optimal_j}
J(\pi_{\theta})= \mathbb{E}_{\pi_{\theta}} \left [ \sum_{t=0}^{\infty} \gamma^{t}r_t \right]
\end{equation}

This objective function is created to develop a policy for the aerial robot to learn the hovering task in the RotorPy \cite{spencer_rotorpy_2023} simulation environment with PPO algorithm \cite{schulman_ppo_2017}. 

\subsection{Reward Function Design}

The hovering task can be formulated as an optimization problem that aims to minimize the position error, ensure stability, and reduce the cost of control actions. In the reward function designed for this task, the variables named state $\bm{s}$ mentioned above are used as observation space. The reward function is designed as follows:

\begin{equation}\label{r_total}
    \mathbb{R}_{\mathrm{total}} = \mathbb{R}_{\mathrm{dis}} + \mathbb{R}_{\mathrm{vel}} + \mathbb{R}_{\mathrm{action}} 
\end{equation}

The reward function consists of distance to the target location, velocity and action components. The reward components are detailed as follows:

\begin{equation}
\begin{aligned}
    \mathbb{R}_{\mathrm{dis}} &= -w_{p}\|\mathbf{p}\|, \\
    \mathbb{R}_{\mathrm{vel}} &= -w_{v}\|\mathbf{v}\|, \\
    \mathbb{R}_{\mathrm{action}} &= -w_{a}\|\mathbf{a}\|
\end{aligned}
\end{equation}

\noindent where $\mathbb{R}_{\mathrm{dis}}$ is a distance-based component that penalizes the position of the aerial robot with respect to the Euclidean distance from the origin. $\mathbb{R}_{\mathrm{vel}}$ is a component that is used to ensure that the aerial robot exhibits stable velocity behaviour and penalizes it according to the velocity norm. $\mathbb{R}_{\mathrm{action}}$ is a component of the reward function used to penalize control effort. All components of the reward function are scaled by the hyperparameter $[w_p, w_v, w_a]$. The parameters specified in \cref{tableReward} were used in the experimental studies.


\begin{table}[htbp!]
\centering
\caption{\label{tableReward}Hyperparameter of the reward function for the training.}
\begin{tabular}{cc}
\hline
\textbf{Parameter} & \textbf{Value} \\ \hline
\rowcolor[HTML]{f2f2f2} 
$w_p$              & 1.0            \\
$w_v$              & 0.5            \\
\rowcolor[HTML]{f2f2f2} 
$w_u$              & 1e-5           \\ \hline
\end{tabular}%
\end{table}

\subsection{Retrospective Cost Mechanism}
\begin{figure}[t]
      \centering
      \includegraphics[scale=.35]{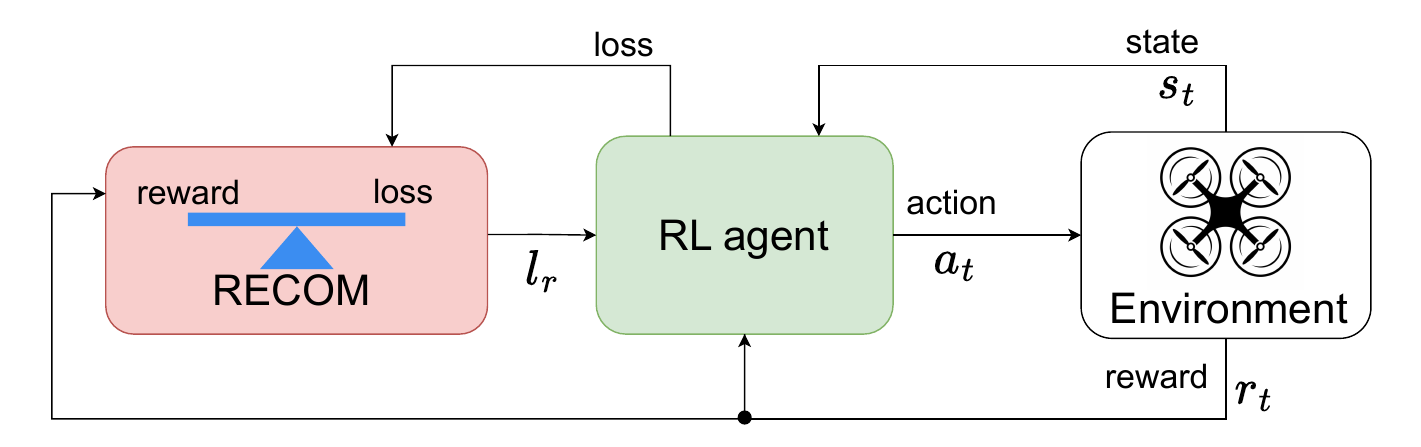}
      \caption{Our RECOM framework uses loss and reward information during training. RECOM is designed to adapt to non-stationary environments in RL. It calculates a dynamically updated learning rate that balances rewards and losses.}
      \label{recom_fig}
   \end{figure}

RECOM is used to adapt the long-term learning process. It has been a preferred approach, especially with adaptive controllers \cite{dennis_rcac_2010}. 
RECOM has been proposed to improve decision-making, prevent loss of plasticity, and balance rewards and losses during long-term training in RL. The approach used in RECOM is shown in \cref{recom_fig}.
We present the formulation of the RECOM as follows:


\begin{equation}
    \mathbb{C}_{\mathrm{ret}} = \frac{1}{T} \sum_{t=N-T+1}^{N} (-\mathbb{R}[t]+\mathbb{L}[t])
\end{equation}
\noindent where $\mathbb{R}[t]$ and $\mathbb{L}[t]$ represent rewards and losses, $T$ is the retrospective window and $N$ is the length of the array that includes rewards and losses. In this formulation, the retrospective window corresponds to the last $T$ episode. Here, the mean of rewards and losses received by the last $T$ episode is calculated. The previous cost is calculated as follows:


\begin{equation}
    \mathbb{C}_{\mathrm{prev}} = \frac{1}{T} \sum_{t=N-T}^{N-1} (-\mathbb{R}_{\mathrm{prev}}[t]+\mathbb{L}_{\mathrm{prev}}[t])
\end{equation}
\noindent where $\mathbb{R}_{\mathrm{prev}}[t]$ and $\mathbb{L}_{\mathrm{prev}}[t]$ represent rewards and losses for the previous retrospective window. According to the latest and previous costs, the cost gradient is defined as

\begin{equation}
    \mathbb{G}_{\mathrm{cost}} =  \mathbb{C}_{\mathrm{ret}} - \mathbb{C}_{\mathrm{prev}}
\end{equation}

After the cost gradient value is calculated for each retrospective window, the RECOM can update the learning rate in the PPO as

\begin{equation}
    \mathbb{L}_{r} \leftarrow  \mathbb{L}_{r} - \mathbb{R}_{\mathrm{gain}} \times \mathbb{G}_{\mathrm{cost}}
\end{equation}
where $\mathbb{R}_{\mathrm{gain}}$ is a parameter that comes with the RECOM mechanism. This hyperparameter has been used as a value $[5e-6]$. According to the RECOM approach, the learning rate is updated in each 40K training step. 

\subsection{Training and Implementation Details}
We selected the PPO algorithm from the Stable Baselines3 (SB3) framework \cite{raffin_sb3_2021} for both its robust policy optimization and its efficient handling of high-dimensional continuous action spaces. A policy network consisting of a two-layer multilayer perceptron with 64 neurons per layer was created for policy training. The activation function is $tanh(\cdot)$, and the last layer of the actor net outputs a 4-dimensional vector. The policy network was trained for a total of about 20 million timesteps. We used Adam optimizer \cite{diederik_adam_2015} with a dynamic learning rate $\mathbb{L}_{r}$ update by RECOM and a batch size of 64. In L2 regularization with PPO and RECOM with L2 PPO, we added L2 regularization to prevent overfitting and improved generalization by penalizing large weights in the policy network. We investigated the behaviour of RECOM against loss of plasticity in the aerial robot both without wind disturbance and with wind disturbance for 20 million timesteps. In the wind perturbation tests, the intensity of the wind perturbation varied every 2 million timesteps. The components and physical parameters of the aerial robot used to develop a policy for the hovering task with RL are shown in \cref{tableRobot}.

\begin{table}[b!]
\centering
\caption{\label{tableRobot}Overview of the physical parameters of the aerial robot used in the simulation environment.}
\begin{tabular}{lc}
\hline 
\textbf{Parameter (unit)} & \textbf{Crazyflie Platform}\\ 
\hline 
\rowcolor[HTML]{f2f2f2} 
Mass, m (\unit{kg})&0.03\\
Inertia, J (\unit{kg.m^2})&[1.43e-5, 1.43e-5, 2.89e-5]\\
\rowcolor[HTML]{f2f2f2} 
Arm length (\unit{m}) & 0.043\\
Thrust-to-Weight-Ratio & 1.95 \\
\rowcolor[HTML]{f2f2f2} 
Maximum thrust (\unit{\newton}) & 0.575 \\
Motor time constant (\unit{s}) & 0.05\\
\hline 
\end{tabular}
\end{table}

\subsection{Simulation Environment Setup}

We used the RotorPy simulation environment, which provides various aerodynamic wrenches \cite{lee_se3_2010}, and was based on Python, both lightweight and easy to install with few dependencies or requirements. This setup provided an efficient aerial robot simulation with customizable dynamics and control algorithms, in particular, to develop and test different RL algorithms. We had done all of the simulations and RL training on a laptop equipped with an Intel\textregistered  Alder Lake Core\texttrademark  i7-12700H and NVIDIA\textregistered GeForce  RTX4060\texttrademark (8GB GDDR6) GPU and 32 GB DDR4 RAM (2x16GB, 3200MHz).

\section{Results and Discussions}
\subsection{Simulation Training and Evaluation}

\begin{figure}[t!]
      \centering
      \includegraphics[width=1\linewidth]{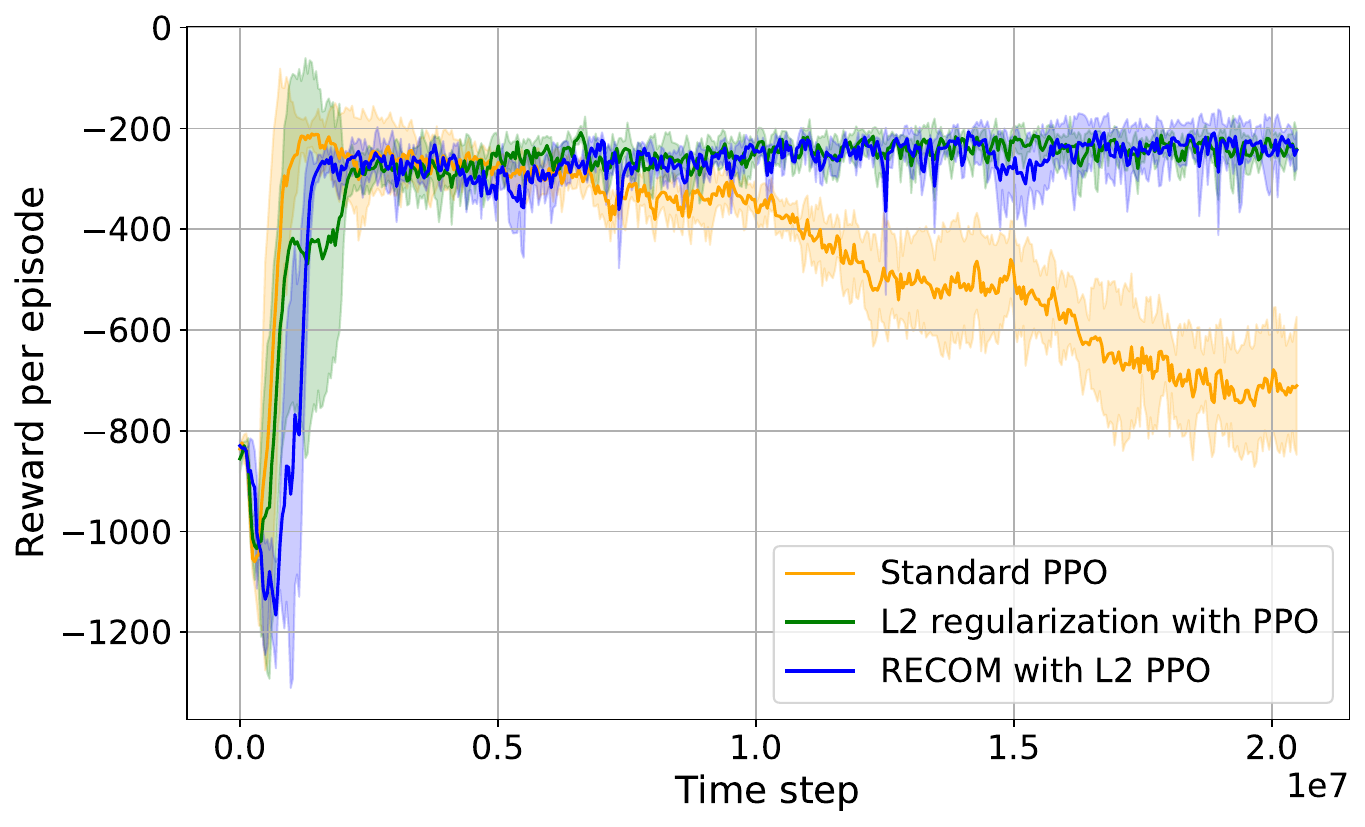}
      \caption{Comparison of different reinforcement learning agents in training performance with wind disturbance.}
      \label{wind_result}
   \end{figure}

We evaluated the hovering flight control policy learned by the PPO agent with wind perturbations. To investigate the loss of plasticity and maintaining plasticity strategies, we trained three different RL agents, including standard PPO, L2 regularization with PPO and RECOM with L2 PPO. In \cref{wind_result}, we have shown the average rewards for the three agents, averaged over 5 different experiments performed at 20 million timesteps.

\begin{figure}[t!]
      \centering
      \includegraphics[width=1\linewidth]{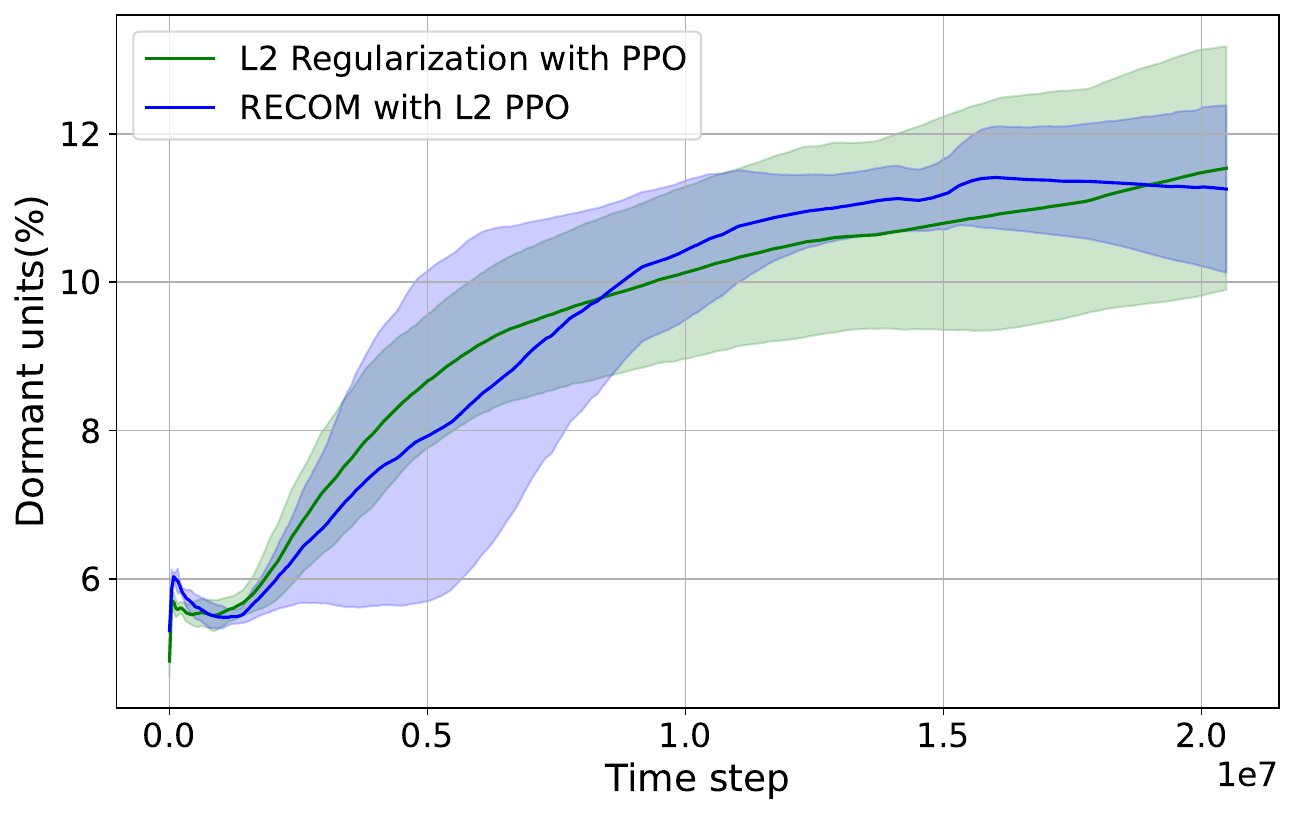}
      \caption{Change of dormant units in the policy network during training under the wind disturbance.}
      \label{wind_dormant}
   \end{figure}

During the 20 million training, the wind disturbance value was updated every 2 million timesteps to be \qtylist{3.0; 2.0; 2.5; 1.5; 2.5}{\metre\per\second}, respectively, to demonstrate the loss of plasticity. The baseline agent, the standard PPO, has a policy crash after 10 million timesteps compared to other agents. Additionally, the standard PPO has received lower rewards. 
The reason for this situation is the increase in dormant units in the standard PPO network during long-term training, a problem analyzed in detail in \cite{dohare_loss_2024}.
In contrast, RL agents running the PPO algorithm regulated by L2 regularization and RECOM achieved higher rewards. Whether the mechanisms developed for RL produce dormant units in the network during long-term training should also be considered as an evaluation criterion to be checked. For this reason, dormant units ratio in the policy network is shown in \cref{wind_dormant}.

During the policy training for long periods, it was observed that the L2 and RECOM approaches produced successful results. When we analysed the change in dormant units with the L2 and RECOM approaches, we observed similar changes. However, the L2 regularization with PPO was shown to have 2.42\% more dormant units than RECOM with L2 PPO. This observation led us to investigate the phenomena of dormant units in different conditions. 

\begin{figure}[b!]
      \centering
      \includegraphics[width=1\linewidth]{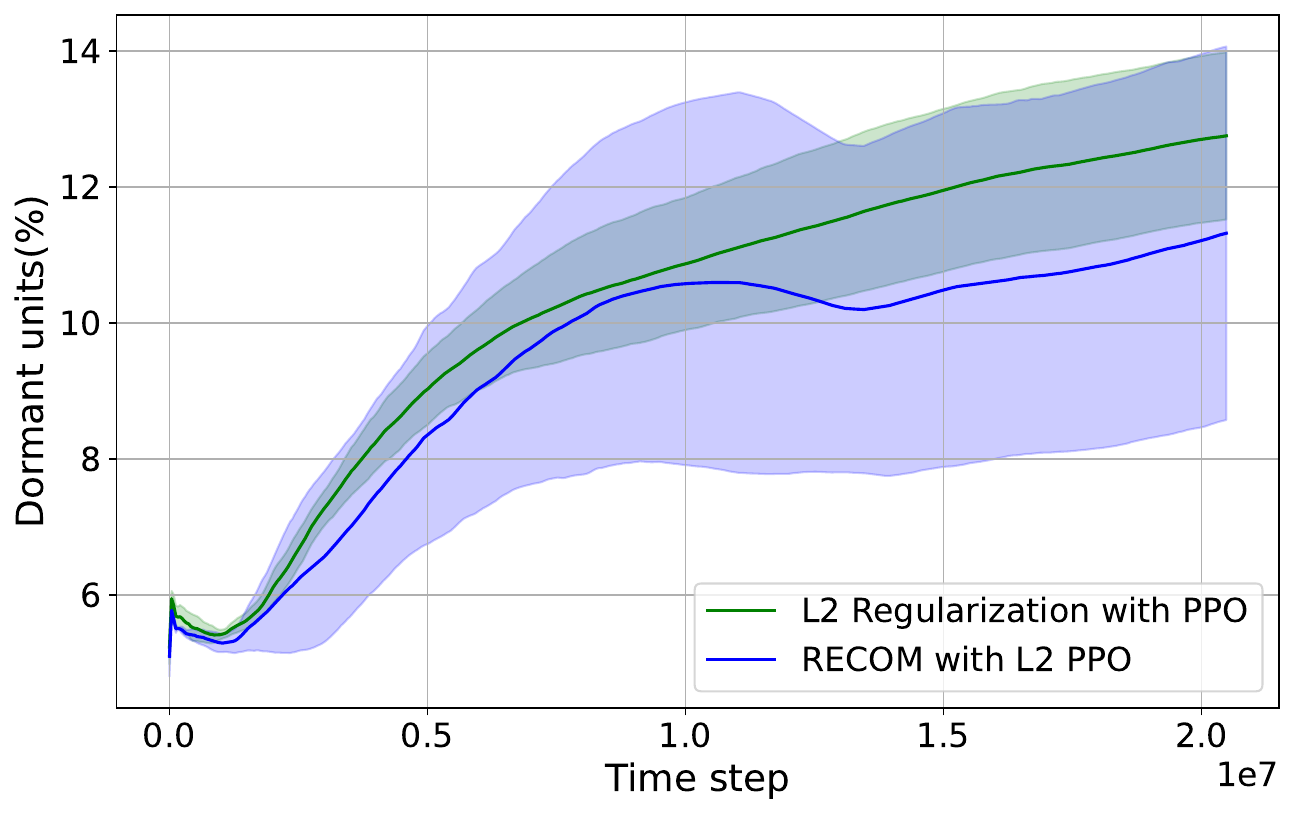}
      \caption{Change of dormant units in the policy network during training without the wind disturbance.}
      \label{without_dormant}
   \end{figure}

As a further investigation, it was tested whether the loss of plasticity occurs without wind perturbation. These tests were also conducted to determine if the results obtained under wind disturbance conditions would be maintained in a stationary environment. 
%
%
During the long-term training, both in L2 regularization with PPO and RECOM with L2 PPO, the agent has developed a policy to maximize the required reward. Dormant units ratio in the policy network was shown in \cref{without_dormant}.

From the results of long-term training without wind disturbance, it was observed that the rate of dormant units in the network using the RECOM mechanism was lower. 
As a result of all the tests, RECOM with the L2 PPO approach has developed a policy to maximize reward and has achieved a successful result of 11.29\% less dormant units in the policy network than L2 regularization at 20 million timesteps. The difference between these RL agents in terms of dormant units was observed to be higher than those in stationary environments. This was considered to be because the policy network tends to maintain the established policy rather than change it in stationary environments.

\subsection{Validation of Learned Policies}

\begin{figure}[t!]
      \centering
      \includegraphics[width=1\linewidth]{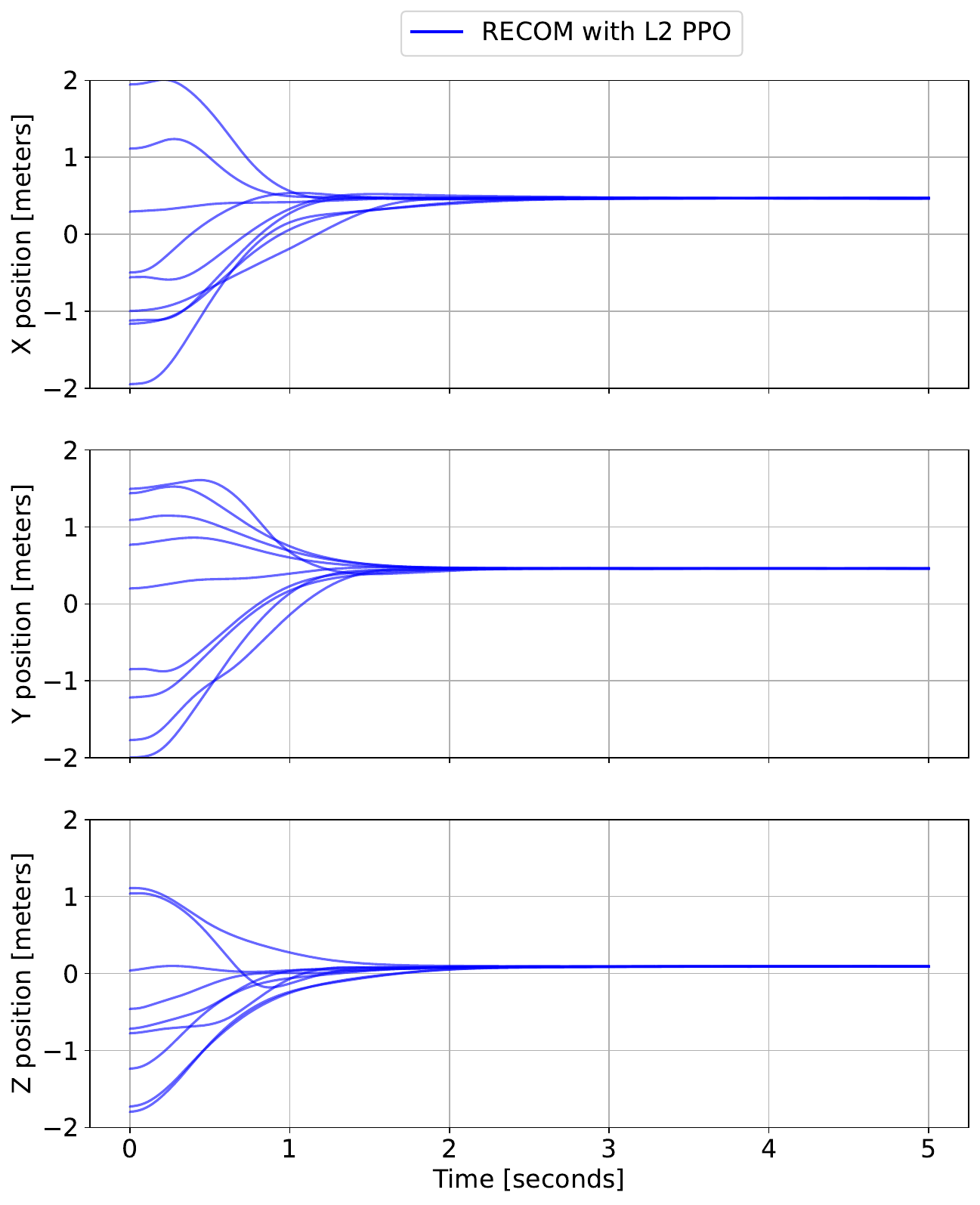}
      \caption{Evaluating the performance of different agents under the wind disturbance.}
      \label{eval_rl_wind}
   \end{figure}

We evaluated the hovering control policies learned by the RL agents under 3 \unit{\meter\per\second} wind disturbance. Additionally, we compared their performance with standard PPO, L2 regularization with PPO, and RECOM with L2 PPO in the RotorPy. The result of RECOM with L2 PPO agent evaluation is shown in \cref{eval_rl_wind}.
   
In total, we applied 10 evaluations with different initial conditions of RL agents for each control policy. The initial positions were assigned to random positions between \qtyrange{-2}{2}{\metre}. All agents were asked to move to reference position 0 on all axes for the hovering task. We used the policy checkpoint at 20 million timesteps.
In these experiments, we used the RECOM with the L2 PPO policy that produced the highest reward during the simulation. It was observed that 90\% success rate of the RL agents positioned themselves in the reference position given for the hovering task. RECOM with L2 PPO policy showed the ability to achieve hovering tasks in long-term training settings with wind perturbation.





\begin{table}[htbp!]
\centering
\caption{\label{tableExp} Simulation results: MSE error metrics comparison for three policies.}
\resizebox{\columnwidth}{!}{%
\begin{tabular}{lccccc@{}}
\toprule
\multicolumn{1}{l}{}                                      & \textbf{Success Rate $\uparrow$} & \multicolumn{3}{c}{\textbf{MSE of Position Error $\downarrow$}} \\
\multicolumn{1}{l}{\multirow{-2}{*}{\textbf{Control Policy}}} & \textbf{(\%)}         & \textbf{X (m)}  & \textbf{Y (m)}  & \textbf{Z (m)} \\ \midrule
\rowcolor[HTML]{f2f2f2} 
Standard PPO                                              & 30                    & 0.60            & 1.27            & 0.25           \\
L2 regularization with PPO                                & 88                    & 0.38            & 0.49            & \textbf{0.09}  \\
\rowcolor[HTML]{f2f2f2} 
RECOM with L2 PPO                                         & \textbf{90}           & \textbf{0.32}   & \textbf{0.35}   & 0.10           \\ \bottomrule
\end{tabular}%
}
\end{table}

\begin{figure}[t!]
		\centering
		\begin{overpic}[scale=0.38,trim=35 5 5 5]{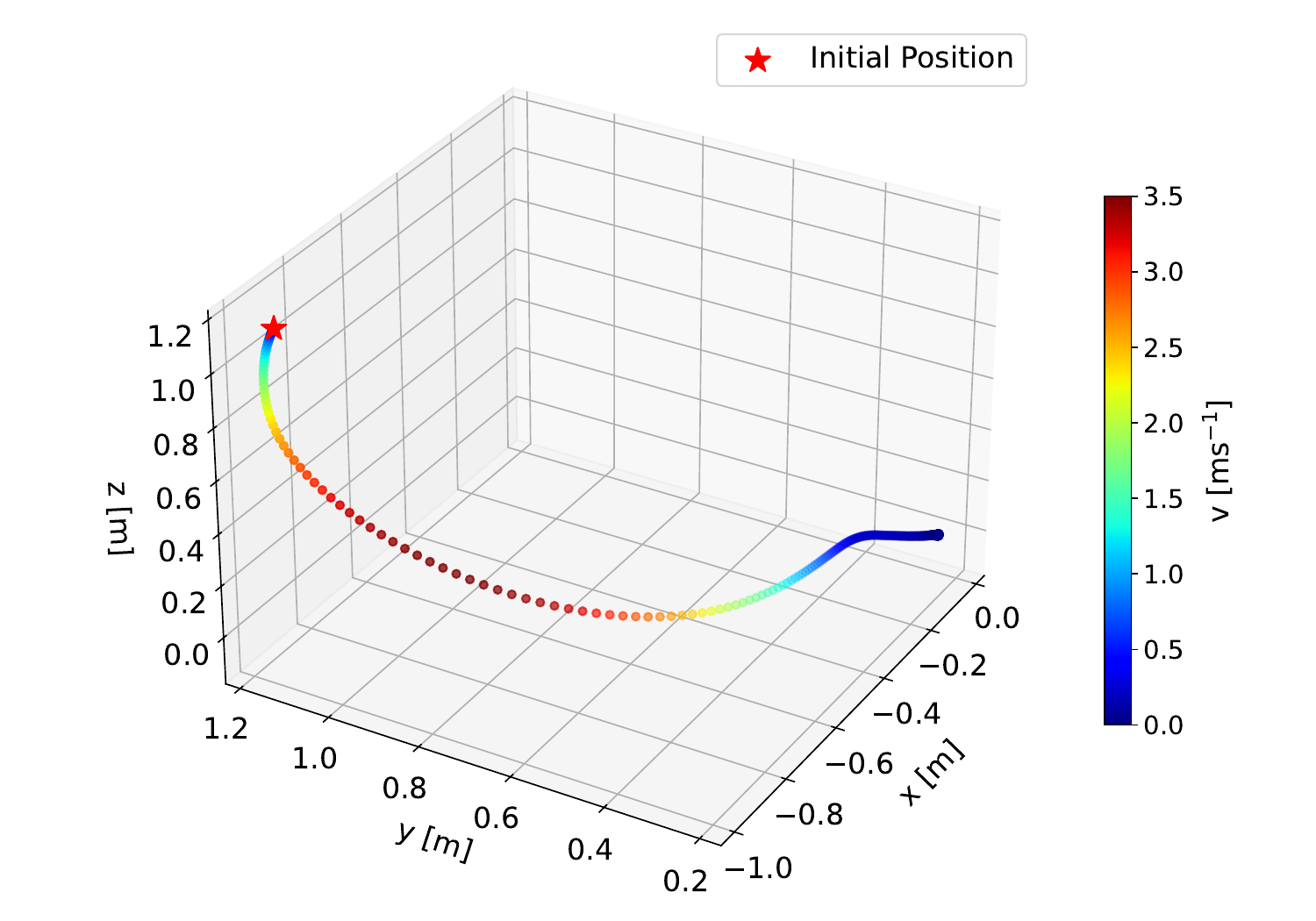}            
		\end{overpic}
\caption{ Illustration of the performance of an RL agent under the wind disturbance.}\label{3d_hover_eval}
\end{figure}

In \cref{tableExp}, we used the mean square error (MSE) metric to evaluate the position error of different RL agents. 
The standard PPO achieved a success rate of only 30\% in a non-stationary environment setting, with relatively large tracking errors in three axes due to policy collapse. In contrast, L2 regularization with PPO substantially improved the success rate to 88\% while also significantly reducing the overall position tracking error compared with standard PPO. Our proposed RECOM with L2 PPO method demonstrated the highest success rate at 90\% and the smallest horizontal errors.
There are also linear MPC approaches with sub-centimeter performance, as shown in \cite{davide_2022_mpc_rl}. However, the goal of this paper is not only to propose a new controller for aerial robotics applications, but also to show that a mechanism that balances rewards and losses, inspired by neuroscience and cognitive science, can evolve a policy for hovering tasks without policy collapse in long term training with non-stationary wind environment. An example of successful hovering under wind disturbance was visualized in \cref{3d_hover_eval}.

According to the results, we observed that the aerial robot reached 3.5 \unit{\meter\per\second} during the hovering task. Then the RL agent was observed to stay hovering, reaching the position setpoint on each axis.

\subsection{Discussions}
Our experiments have carefully evaluated the proposed RECOM with L2 PPO framework, focusing on the long-term training performance of the aerial robot for the hovering task. According to the results, we show that the RECOM with L2 PPO both prevents policy collapse compared to standard PPO and organizes dormant units better than L2 regularization with PPO. While the standard PPO experienced a dramatic policy collapse after 10 million timesteps, RECOM with L2 PPO continued to attempt to maximize reward in the long-term training of the hovering task. 
In addition to achieve these gains, it showed that under the wind disturbance a validation performance of 90\% according to the results in the RotorPy simulation environment. To better understand the contribution of rewards and losses to RL and their impact on policy stability, we considered insights from cognitive science and neuroscience.

Rewards are the key driver for agent learning in RL which serves as positive feedback to reinforce desirable actions. In the human brain, this process corresponds closely to the dopaminergic system \cite{dopamine}, particularly in structures such as the basal ganglia \cite{basal_ganglia} and ventral striatum \cite{ventra_striatum}. Cognitive neuroscience considers a “reward” to be any stimulus that the brain associates with desirable outcomes, which in turn triggers increases in dopaminergic activity \cite{cognitive_reward}. According to principles of Hebbian learning \cite{hebb_theory}, neurons that are coactivated by rewarding outcomes tend to strengthen their connections, thereby reinforcing the behaviours that led to the reward. This mechanism is also linked to a “primacy bias", in which early high rewards can overshadow subsequent events, making it harder to retain plasticity later in training if the learning rate remains fixed.

In contrast, losses or punishments act as negative feedback. From a neuroscience standpoint, losses activate structures such as the amygdala and anterior cingulate cortex, which are critical for error detection and avoidance learning \cite{losses_ACC_2006}. Similar to how reward encourages a neural trace to be strengthened, losses can drive long-term depression in synaptic efficacy \cite{sheng_long_term_2014}, effectively discouraging repeated engagement in maladaptive behaviours. In cognitive and neuroscientific terms, loss-driven signals often require more careful processing to avoid underreacting or overcorrecting, thus supporting a more gradual, corrective component in decision-making \cite{botvinick_reinforcement_2019}.


In standard RL formulations, these two signals, reward maximization and loss minimization, are usually handled indirectly (e.g., through a single scalar reward plus a training loss objective). However, by explicitly balancing the role of reward and loss in one retrospective cost term, RECOM is designed to emulate a richer feedback loop by evoking how biological systems adapt behaviour. Specifically, RECOM’s retrospective cost $\mathbb{C}_{\mathrm{ret}}$ combines the negative of cumulative reward $-\mathbb{R}[t]$ with the training loss $\mathbb{L}[t]$ into a single scalar. Therefore, it models the interplay between dopaminergic “go” signals (reward) and cortical/subcortical “caution” or “error” signals (loss). This unified feedback drives an adaptive update to the PPO learning rate, thereby aiding to preserve the agent’s plasticity over long horizons and contributing to preventing policy collapse. By periodically re-evaluating the learning rate based on recent rewards and losses, the RECOM allows the agent to both avoid policy collapse and reactivate dormant units more successfully and to adapt to changing environmental conditions throughout long-term training. Additionally, the RECOM update acts much like a negative feedback control loop, lowering the learning rate if the recent cost rises and increasing it if the cost drops. This aligns with how neural circuits modulate synaptic plasticity in response to aggregated outcomes: reward signals can boost learning rates, while substantial errors can dampen them to maintain stability.

However, it should be noted that our framework has some limitations and similar results cannot be produced for all PPO variations. We based our findings on the SB3 framework, the PPO default settings used in our experiments, and the observed policy collapse in standard PPO. This may not be true for all PPO variations or hyperparameter configurations. It should also be noted that while RECOM with L2 performs well in hovering tasks, it may not be an effective solution for agile flight or trajectory tracking tasks. It is also important to note that more research is needed to see how our approach works with different RL algorithms, how well it performs in real-world experiments, and how it works on different aerial robot tasks.
\section{Conclusion}

In this work, we proposed a retrospective cost mechanism (RECOM) to balance rewards and losses specifically focusing on the aerial robot for hovering tasks in RotorPy with non-stationary environment scenarios. 
By integrating RECOM with L2 PPO, we successfully prevent policy collapse during long-term training, achieving 11.29\% fewer dormant units in the policy network compared to standard PPO with L2 regularization after 20 million timesteps. 
The experimental results demonstrated that rewards and losses could be balanced with a retrospective mechanism inspired by the perspective of neuroscience and cognitive science. 

Future work will focus on extending this framework to real-world experiments in non-stationary environments, considering the sim-to-real gap problem.

\addtolength{\textheight}{-13cm}




\section*{Acknowledgement}

This research was partially supported by seedcorn funds from Civil, Aerospace and Design Engineering, Isambard AI, and Bristol Digital Futures Institute at the University of Bristol.



\bibliographystyle{IEEEtran}
\bibliography{ref}

\end{document}